\documentclass[11pt]{article}

\usepackage[preprint]{acl}

\usepackage{times}
\usepackage{latexsym}
\usepackage[T1]{fontenc}
\usepackage[utf8]{inputenc}
\usepackage[no-math]{fontspec}
\newfontfamily\khmerfont[Script=Khmer,Scale=0.95]{NotoSerifKhmer-Regular.ttf}

\newcommand{\khmer}[1]{{\khmerfont\XeTeXlinebreaklocale "km" #1}}
\newcommand{\lat}[1]{{\fontencoding{T1}\fontfamily{ptm}\selectfont #1}}
\usepackage{microtype}
\usepackage{inconsolata}
\usepackage{graphicx}

\usepackage{amsmath}
\usepackage{amssymb}
\usepackage{booktabs}
\usepackage{array}
\usepackage{subcaption}
\usepackage{tikz}
\usetikzlibrary{arrows.meta,positioning,calc}
\definecolor{accent}{RGB}{31,94,166}
\newlength{\ovw}
\newcommand{\fmk}[1]{\tikz[baseline=(c.base)]\node[circle, fill=accent, text=white, font=\scriptsize\bfseries, inner sep=0pt, minimum size=9pt](c){#1};}

\title{Khmer PFGen-Bench: Building a Judge-Free Benchmark for Open-Ended Generation in a Low-Resource Language}

\author{Mengsay Loem, Kentaro Imajo, Ryota Ozaki, Masanori Hirano\\
  Preferred Networks, Inc.\\
  \textnormal{\small\{mengsay, imos, ryota55ozaki\}@preferred.jp, research@mhirano.jp}}

\begin{document}
\maketitle

\begin{abstract}
Benchmarks for open-ended generation often rely on large language model (LLM) judges, whose reliability in low-resource languages is uncertain.
We introduce Khmer PFGen-Bench, to our knowledge the first judge-free benchmark for multi-sentence answers to questions composed directly in Khmer, a low-resource language.
Khmer PFGen-Bench adapts the scorer of PFGen-Bench, a benchmark for Japanese that scores responses with keyword rules and n-gram statistics from LLM-generated references.
This scoring requires references that are both valid and varied, which LLMs may not produce in Khmer, so we examine how to build such references.
We find that post-trained models can give more valid but less varied references than their base variants, and that mixing model families increases variety for the same number of references.
Using references from post-trained models in nine families, the benchmark yields model rankings that are stable across reference sets and consistent with a native speaker's ratings.
These findings offer guidance for building judge-free benchmarks in other low-resource languages.
\end{abstract}

\section{Introduction}

\begin{figure*}[t]
  \centering
  \begin{minipage}[c]{\ovw}
  \centering
  \begin{tikzpicture}[
    font=\small,
    box/.style={draw=black!35, rounded corners=2pt, align=center, inner sep=3pt, text width=#1},
    arr/.style={-{Stealth[length=4pt]}, draw=black!55, line width=0.6pt},
    lab/.style={font=\footnotesize\itshape, text=black!65, fill=white, inner sep=1pt},
    mk/.style={circle, fill=accent, text=white, font=\scriptsize\bfseries, inner sep=0pt, minimum size=9pt},
  ]
    \node[box=0.95\ovw] (q) {%
      \textbf{50 questions written directly in Khmer}\\
      {\footnotesize 13 categories from the Cambodian curriculum}\\[2pt]
      \khmer{ចូររៀបរាប់ពីអត្ថប្រយោជន៍នៃក្រមាខ្មែរ។}\\
      {\footnotesize\textit{Describe the uses of the Khmer \emph{krama} (scarf).}}};
    \node[box=0.44\ovw, anchor=north east, minimum height=1.75cm]
      (g) at ([xshift=-0.015\ovw, yshift=-6mm]q.south) {%
      \textbf{Reference generators}\\
      {\footnotesize 9 model families\\ 29,321 $\rightarrow$ 18,274\\ after filtering}};
    \node[box=0.44\ovw, anchor=north west, minimum height=1.75cm]
      (m) at ([xshift=0.015\ovw, yshift=-6mm]q.south) {%
      \textbf{Evaluated models}\\
      {\footnotesize 19 models, including\\ the 9 generators}};
    \node[box=0.95\ovw, anchor=north, fill=accent!6]
      (s) at ([yshift=-6mm]$(g.south)!0.5!(m.south)$) {%
      \textbf{Judge-free scoring} {\footnotesize (PFGen-Bench scorer)}\\
      {\footnotesize F: n-gram fluency \quad T: 3-gram support\\ H: question-specific keyword rules}};
    \node[box=0.95\ovw, anchor=north, dashed]
      (h) at ([yshift=-6mm]s.south) {%
      \textbf{Native-speaker ratings} {\footnotesize (validation only)}\\
      {\footnotesize 240 responses from 6 models}};
    \draw[arr] (q.south -| g.north) -- (g.north);
    \draw[arr] (q.south -| m.north) -- (m.north);
    \draw[arr] (g.south) -- node[lab] {references} (g.south |- s.north);
    \draw[arr] (m.south) -- node[lab] {responses} (m.south |- s.north);
    \draw[arr, dashed] (s.south) -- node[lab, right=1pt, fill=none] {compared} (h.north);
    \node[mk] at (g.north west) {1};
    \node[mk] at ([xshift=11pt]g.north west) {2};
    \node[mk] at (s.north west) {3};
    \node[mk] at (h.north west) {4};
  \end{tikzpicture}
  \end{minipage}\hfill
  \begin{minipage}[c]{0.5\textwidth}
  \footnotesize
  \textbf{What this paper tests}\par\smallskip
  \begin{tabular}{@{}c@{\hspace{5pt}}>{\raggedright\arraybackslash}p{0.92\linewidth}@{}}
    \fmk{1} & \textcolor{accent}{\textbf{Do generators produce valid references?}} (\S\ref{sec:yield})\newline
              \textit{Varied:} base vs.\ post-trained variants; Khmer vs.\ Japanese.\newline
              \textit{Finding:} estimated valid-reference yield is 0--30\% in Khmer and 29--68\% in Japanese. \\[4pt]
    \fmk{2} & \textcolor{accent}{\textbf{Do the references cover varied expressions?}} (\S\ref{sec:coverage}--\ref{sec:mixing})\newline
              \textit{Varied:} model variant; one to eight families at a fixed reference count.\newline
              \textit{Finding:} post-trained variants have lower coverage; mixing families raises it. \\[4pt]
    \fmk{3} & \textcolor{accent}{\textbf{Are rankings stable across reference sets?}} (\S\ref{sec:allocation}--\ref{sec:stability})\newline
              \textit{Varied:} reference count, family subsets, self-inclusion, family replacement.\newline
              \textit{Finding:} $\rho \ge 0.95$, but a generator's own references raise its score. \\[4pt]
    \fmk{4} & \textcolor{accent}{\textbf{Do scores agree with a native speaker?}} (\S\ref{sec:human})\newline
              \textit{Compared:} scorer outputs and ratings for 240 responses.\newline
              \textit{Finding:} model ranking on rated responses $\rho = 0.94$; within-question comparisons among frontier responses are near chance. \\
  \end{tabular}
  \end{minipage}
  \caption{Khmer PFGen-Bench (left) and the conditions behind its scores that this paper tests (right). The PFGen-Bench scorer compares each response with separately generated references, without an LLM judge; numbers link each condition to the component it concerns.}
  \label{fig:overview}
\end{figure*}

Large language models (LLMs) are widely used as judges of open-ended generation \citep{zheng-etal-2023-judging}, but whether an LLM judge is reliable in a low-resource language has to be checked in that language.
A model's ability to generate target-language text does not by itself establish reliable judging ability.
Multilingual studies report biased or inconsistent judgments in low-resource languages, highlighting the need for validation against native-speaker assessments \citep{hada-etal-2024-large,son-etal-2024-mmeval}.
Concerns about judgment reliability motivate evaluation methods that do not require an LLM judge at scoring time.

These concerns apply to Khmer, the official language of Cambodia and a low-resource language.
Existing benchmarks that include Khmer translate their items from other languages \citep{han-etal-2026-deepsea}, score only short answers by exact match \citep{hupkes-bogoychev-2025-multiloko}, or score open-ended answers with an LLM judge \citep{nguyen-etal-2024-seallms}.

To address these limitations, we present Khmer PFGen-Bench, a judge-free benchmark that uses questions composed directly in Khmer by a native speaker on Cambodian school-curriculum topics and scores multi-sentence answers against reference answers.
To our knowledge, it is the first Khmer benchmark to combine these three properties.
Its scoring method comes from PFGen-Bench \citep{imajo-etal-2025-judge}, which was developed for Japanese and combines question-specific keyword rules with n-gram statistics from reference answers generated separately from the evaluated responses.

The main challenge in adapting PFGen-Bench to Khmer is constructing the reference answers; the scoring procedure itself carries over with language-specific text processing and parameter settings.
PFGen-Bench treats the reference answers for each question as an approximation of the distribution of desirable answers, which presumes generators that produce valid and varied answers in the target language.
This premise is plausible for Japanese but cannot be taken for granted in a low-resource language such as Khmer.
We therefore test four conditions behind the scores (Figure~\ref{fig:overview}): (1) whether generators produce valid references, (2) whether the references cover varied expressions, (3) whether model rankings are stable across reference sets, and (4) whether the scores agree with a native speaker's judgments.

Conditions (1) and (2) matter because incorrect references distort the n-gram statistics and narrow references may penalize valid responses phrased differently, as studies of reference diversification suggest \citep{tang-etal-2024-metrics,zeng-etal-2024-towards}; conditions (3) and (4) determine whether the scores support model comparison.
We measure (1) by \emph{valid-reference yield}, the estimated share of generated candidates that are retained and valid, and (2) by \emph{expressive coverage}, the range of answer expressions represented in the references.
Since instruction tuning and preference training can affect instruction following and output diversity \citep{shaham-etal-2024-multilingual,kirk-etal-2024-understanding}, we compare these measures between base and post-trained variants in Khmer and Japanese, and across mixtures of model families.
For (3) and (4), we vary the references and compare the benchmark ranking with a native speaker's ratings.

Khmer PFGen-Bench pairs 50 questions with 18,274 references from post-trained generators in nine model families and scores each model without that model's own references.
The analyses support these settings: post-trained variants can have higher yield but lower coverage than base variants, combining families raises coverage at a fixed reference count, and a generator's own references raise its score.
Model rankings remain highly correlated across reference sets and agree with a native speaker's ratings at the model level.

This approach, which assesses generators for both yield and coverage and excludes self-references in scoring, can guide the construction of judge-free benchmarks in other low-resource languages.
We plan to release the benchmark data and scoring code.

\section{Related Work}

\paragraph{LLM judges in multilingual evaluation.}
\citet{zheng-etal-2023-judging} document position and verbosity biases in LLM judgments and report that some judges favor their own responses, although the evidence for this self-enhancement bias is inconclusive.
In multilingual evaluation, \citet{hada-etal-2024-large} report inflated GPT-4 scores and emphasize calibration against native-speaker judgments.
MM-Eval further identifies inconsistent and unfair assessments in lower-resource languages \citep{son-etal-2024-mmeval}.
These reliability concerns motivate a judge-free scorer, and the emphasis on target-language validation motivates our comparison of the Khmer scorer with a native speaker's ratings.

\paragraph{Southeast Asian and Khmer benchmarks.}
SeaExam converts official school exam questions from Indonesia, Thailand, and Vietnam into multiple-choice items, and SeaBench scores open-ended, multi-turn responses in the same languages with GPT-4o as the judge \citep{liu-etal-2025-seaexam}.
SEA-HELM evaluates general NLP tasks, instruction following and multi-turn chat, linguistic and cultural capabilities, and safety in Filipino, Indonesian, Tamil, Thai, and Vietnamese \citep{susanto-etal-2025-seahelm}.
Among resources that include Khmer, the current SEA-HELM release adds Khmer tasks for multiple-choice reading comprehension, machine translation, culturally grounded natural language inference, and instruction following and multi-turn chat adapted from IFEval and MT-Bench.\footnote{\url{https://github.com/aisingapore/SEA-HELM}, accessed November 2026.}
Sea-bench, released with SeaLLMs and distinct from SeaBench, combines translated, forum-sourced, and linguist-written instructions in nine languages and scores responses with GPT-4 as the judge \citep{nguyen-etal-2024-seallms}.
MultiLoKo asks native speakers of 31 languages, including Khmer, to write questions about locally relevant topics and scores short answers by exact match \citep{hupkes-bogoychev-2025-multiloko}.
DeepSEA creates its Khmer data by professional translation of existing datasets and scores outputs by accuracy, token-level F1 on extractive answers, or chrF for summarization and MetricX-24 for translation \citep{han-etal-2026-deepsea}.
Khmer PFGen-Bench differs from each of these resources in question origin, answer length, or the use of an LLM judge.

\paragraph{Generated references for evaluation.}
\citet{tang-etal-2024-metrics} expand existing references with LLM-generated paraphrases, while \citet{zeng-etal-2024-towards} generate machine translation references and select candidates for accuracy and diversity.
Reference diversification improves agreement between automatic metrics and human judgments in both studies.
PFGen-Bench \citep{imajo-etal-2025-judge} uses generated answer pools for n-gram and rule-based scoring of open-ended responses and examines score sensitivity to the choice of reference generator.
Khmer PFGen-Bench applies this approach to a low-resource language, separating valid-reference yield from expressive coverage and testing how reference composition and origin affect model rankings.

\paragraph{Generator training and diversity.}
\citet{shaham-etal-2024-multilingual} show that multilingual instruction tuning can improve instruction following in languages seen during pretraining but absent from the tuning data.
For the English-centric models tested by \citet{kew-etal-2024-turning}, cross-lingual transfer is limited by target-language exposure during pretraining.
\citet{kirk-etal-2024-understanding} report lower output diversity after reinforcement learning from human feedback than after supervised fine-tuning on summarization and instruction-following tasks.
The findings on instruction following and diversity motivate our separate assessment of valid-reference yield and expressive coverage in base and post-trained variants.

\section{Benchmark and Study Design}
\label{sec:framework}

\subsection{Questions and Reference Pool}
\label{sec:construction}

The benchmark contains 50 Khmer questions across 13 categories informed by the Cambodian curriculum through secondary school \citep{moeys-2015-curriculum}.
A native Khmer-speaking author wrote the questions and a gold answer for each; the gold answers serve as few-shot examples for other questions, as the basis for keyword rules, and, for ten development questions, as comparison answers (Section~\ref{sec:allocation}).
Table~\ref{tab:examples} shows example questions, and Appendix Tables~\ref{tab:questions-a} and~\ref{tab:questions-b} list all 50.

\begin{table}[t]
  \centering
  \small
  \setlength{\tabcolsep}{3pt}
  \begin{tabular}{@{}l>{\raggedright\arraybackslash}p{0.2\columnwidth}>{\raggedright\arraybackslash}p{0.64\columnwidth}@{}}
    \toprule
    \textbf{ID} & \textbf{Category} & \textbf{Question} \\
    \midrule
    Q01 & Khmer language & \khmer{ចូររៀបរាប់អំពីពាក្យក្លាយក្នុងភាសាខ្មែរ។}\newline
                           \textit{Describe derived words in Khmer.} \\
    \addlinespace
    Q30 & Geography      & \khmer{ហេតុអ្វីបានជាបឹងទន្លេសាបមានទំហំខុសគ្នារវាងរដូវប្រាំងនិងរដូវវស្សា?}\newline
                           \textit{Why does the size of Tonle Sap Lake differ between the dry and rainy seasons?} \\
    \addlinespace
    Q11 & Physics        & \khmer{តើកម្តៅនិងសីតុណ្ហភាពខុសគ្នាដូចម្តេច?}\newline
                           \textit{How do heat and temperature differ?} \\
    \addlinespace
    Q46 & Health         & \khmer{ហេតុអ្វីបានជាភាគល្អិតនិចល \lat{PM2.5} មានគ្រោះថ្នាក់ជាង \lat{PM10}?}\newline
                           \textit{Why is PM2.5 particulate matter more harmful than PM10?} \\
    \bottomrule
  \end{tabular}
  \caption{Example questions in the original Khmer, with English translations for readers.}
  \label{tab:examples}
\end{table}

We use nine post-trained models\footnote{Generators were chosen after a native Khmer speaker's preliminary qualitative review of a small sample of responses from candidate base and post-trained models for correctness and language quality; all nine selected models are post-trained.} from distinct families to construct the main reference pool.
A model family comprises a developer's models and models fine-tuned from them (Appendix Table~\ref{tab:models}).
Four-shot prompting produces 29,321 non-empty, normally completed candidate answers.
Filtering by format, exact duplication, length, and question-specific keyword rules retains 18,274 references.
The benchmark uses all retained references, averaging 365 per question, with counts varying by question and generator.
Rule acceptance does not establish factual correctness.
Appendix~\ref{app:rules} gives the generation settings, selection rules, and reference counts.

\paragraph{Analysis subsets.}
The analyses in Section~\ref{sec:reference-analysis} use 15 Khmer questions and 15 Japanese questions from the original PFGen-Bench; the two sets are not translations of each other.
A separate split assigns ten Khmer questions to development, whose gold answers guide family selection in the allocation study (Section~\ref{sec:allocation}), and 40 to test, which supply the human evaluation (Section~\ref{sec:human}); the leaderboard uses all 50.
The 15-question Khmer subset overlaps both splits; the Khmer reference-generation analyses are not used to tune the scorer or selection rules.

\subsection{Reference Quality Measures}
\label{sec:reference-measures}

\paragraph{Valid-reference yield.}
We estimate the proportion of candidates both retained and valid by combining rule acceptance with audited validity.
For each generator condition, let $n_{\mathrm{cand}}$ and $n_{\mathrm{keep}}$ denote candidate and retained-reference counts.
Of $n_{\mathrm{audit}}$ sampled retained references, $n_{\mathrm{valid}}$ are judged valid through human assessment of factual correctness, completeness, and orthographic well-formedness.
Acceptance $a$, estimated validity $\hat v$, and estimated valid-reference yield $\hat y$ are
\begin{equation}
  a = \frac{n_{\mathrm{keep}}}{n_{\mathrm{cand}}},
  \quad
  \hat v = \frac{n_{\mathrm{valid}}}{n_{\mathrm{audit}}},
  \quad
  \hat y = a\hat v.
  \label{eq:yield}
\end{equation}
Sampling more candidates raises the number of valid references but not their share among retained references: a fraction $1 - \hat v$ of retained references remains invalid, and identifying those references requires native-speaker review.

\paragraph{Expressive coverage.}
We measure expressive coverage against a separate set of answers to the same question, called comparison answers.\footnote{Reference-generation analyses use the official reference set from \citet{imajo-etal-2025-judge} for Japanese and references from held-out model families for Khmer.}
The allocation study uses development gold answers as comparison answers (Section~\ref{sec:allocation}).
We form 3-grams over Japanese characters or Khmer character clusters (KCCs; \citealp{chea-etal-2007-word}).
For question $q$, let $R_q$ be the reference set and $B_q$ the set of comparison answers, with $r \in R_q$ denoting a reference and $b \in B_q$ a comparison answer.
For an answer $x$, either a reference or a comparison answer, $G_3(x)$ is the set of distinct 3-grams in $x$.
The reference 3-gram set $U_q = \bigcup_{r \in R_q} G_3(r)$ contains every 3-gram that occurs in at least one reference.
Coverage is the mean fraction of each comparison answer's distinct 3-grams found in the references:
\begin{equation}
  \mathrm{Cov}_q =
  \frac{1}{|B_q|} \sum_{b \in B_q}
  \frac{|G_3(b) \cap U_q|}{|G_3(b)|}.
  \label{eq:coverage}
\end{equation}

\paragraph{Repetition.}
We adapt the distinct-1 and distinct-2 measures of \citet{li-etal-2016-diversity} to 3-grams.
Distinct-3 is the ratio of distinct 3-grams to all 3-gram occurrences in the reference set:
\begin{equation}
  \mathrm{Distinct\text{-}3}_q =
  \frac{|U_q|}{\sum_{r \in R_q} m_3(r)},
  \label{eq:distinct3}
\end{equation}
where $m_3(r)$ counts 3-gram occurrences in $r$, including repetitions; lower distinct-3 indicates greater n-gram repetition.
Coverage and distinct-3 describe surface-form variation, not semantic diversity or factual correctness.

\subsection{Scoring and Comparison Protocol}
\label{sec:scoring}

We evaluate the nine reference generators and ten models used only for evaluation (non-generators).
Each model answers all 50 questions in ten trials using eight-shot prompting; evaluation responses are generated separately from references and held fixed across reference sets (Appendix~\ref{app:setup}).

We use the three components of the PFGen-Bench scorer, F, T, and H, and follow the definitions of \citet{imajo-etal-2025-judge}, replacing their fixed length and frequency settings with the parameters $\ell_1$, $\ell_0$, and $\tau$.
A response is split into units $C_1, \dots, C_L$: characters in the original Japanese setting and, after normalization, KCCs for Khmer.
Beginning- and end-of-sentence markers are added, giving $C_0, C_1, \dots, C_L, C_{L+1}$.
Let $G^{w}_{i} = \{C_i, C_{i+1}, \dots, C_{i+w-1}\}$ be the $w$-gram that starts at $C_i$, and let $\mathcal{L}^{w}_{G} \in [0,1]$ be the proportion of reference answers for the question that contain the $w$-gram $G$.
Responses longer than $\ell_1$ units are discounted by
\begin{equation}
  d(I) := 1 - \frac{\max(I - \ell_1, 0)}{\ell_0 - \ell_1},
  \label{eq:discount}
\end{equation}
which reaches zero at $\ell_0$ units.
The original setting uses $\ell_1 = 100$ and $\ell_0 = 150$ characters.

\paragraph{Fluency (F).}
F sums the occurrence likelihoods of the $w$-grams in a response for $w = 1, \dots, 10$:
\begin{align}
  F^{\star}_{w} &:= \sum_{i=0}^{L-w+2} \mathcal{L}^{w}_{G^{w}_{i}}, \nonumber \\
  F &:= \sum_{w=1}^{10} d(L)\, F^{\star}_{w}.
  \label{eq:fluency}
\end{align}
F is normalized so that the reference answer set scores 1.0, and the F of a single response can exceed one.

\paragraph{Truthfulness (T).}
T measures how much of a response is supported by frequent reference 3-grams; the name comes from the original benchmark, and T does not verify facts.
Let $\mathcal{L}^{3\star}_{i}$ be the largest $\mathcal{L}^{3}_{G}$ among the 3-grams $G$ that contain $C_i$, and let $\Omega_I \subset \{1, \dots, I\}$ be the indices of non-punctuation units among the first $I$ units.
For a truncation length $I$,
\begin{align}
  T^{\star}_{I} &:= \frac{1}{|\Omega_I|} \sum_{i \in \Omega_I} \frac{\min(\mathcal{L}^{3\star}_{i}, \tau)}{\tau}, \nonumber \\
  T_I &:= d(I)\, T^{\star}_{I},
  \label{eq:truthfulness}
\end{align}
and T takes the best truncation of a long response:
\begin{equation}
  T := \begin{cases}
    T_L & (L \le \ell_1), \\
    \max_{I \in \{\ell_1, \dots, L\}} T_I & (L > \ell_1).
  \end{cases}
  \label{eq:truncation}
\end{equation}
The original setting uses $\tau = 0.005$.
Khmer normalization removes punctuation, so $\Omega_I = \{1, \dots, I\}$ for Khmer.
Units covered only by rare or unseen 3-grams lower T, so content not attested in the reference answers is penalized.

\paragraph{Helpfulness (H).}
H is based on manually defined rules: for each question, keyword elements that a correct answer should contain, combined with AND/OR conditions.
Element $k$ has importance $w_k$, and each element not matched in the first $I$ units multiplies the score by $1 - w_k$:
\begin{equation}
  H_I := d(I) \prod_{k \notin K_I} (1 - w_k),
  \label{eq:helpfulness}
\end{equation}
where $K_I$ is the set of elements matched in $C_1, \dots, C_I$.
H is obtained from $H_I$ with the truncation of Equation~\ref{eq:truncation}.
Of the 216 Khmer elements, 181 have $w_k = 1$, so a response that misses any of them receives $H = 0$; the remaining elements have $w_k = 0.5$.

The benchmark score (Score) is the average $(F+T+H)/3$.
Appendix~\ref{app:selection} describes the Khmer normalization, and Appendix~\ref{app:scoring} gives the Khmer values of $\ell_1$, $\ell_0$, and $\tau$.
Model scores average over questions and trials, including failed generations scored as zero.

For the main leaderboard (Appendix Table~\ref{tab:leaderboard}), F and T are computed separately against references from the five open-weight and the four closed-weight generators and then averaged with equal weight.
Because the reference generators are also evaluated, the leaderboard excludes the evaluated model's own references but keeps references from other models in the same family; Section~\ref{sec:stability} examines the resulting family-level dependence.

\begin{table}[t]
  \centering
  \small
  \setlength{\tabcolsep}{2.5pt}
  \begin{tabular}{@{}llrrr@{}}
    \toprule
    \textbf{Generator} & \textbf{Variant} & \textbf{Acceptance} & \textbf{Validity} & \textbf{Yield} \\
    \midrule
    \multicolumn{5}{@{}l}{\textbf{Japanese}} \\
    Gemma-4-31B   & Base & 80\% & 65\% & 52\% \\
                  & Post & 90\% & 75\% & 68\% \\
    Qwen3-30B-A3B & Base & 65\% & 45\% & 29\% \\
                  & Post & 75\% & 50\% & 38\% \\
    \midrule
    \multicolumn{5}{@{}l}{\textbf{Khmer}} \\
    Gemma-4-31B   & Base & 40\% & 40\% & 16\% \\
                  & Post & 58\% & 52\% & 30\% \\
    Qwen3-30B-A3B & Base &  9\% &  0\% &  0\% \\
                  & Post & 22\% &  8\% &  2\% \\
    \bottomrule
  \end{tabular}
  \caption{Acceptance, audited validity, and estimated valid-reference yield. Post denotes post-trained.}
  \label{tab:yield}
\end{table}

\begin{figure*}[t]
  \centering
  \includegraphics[height=6.2cm]{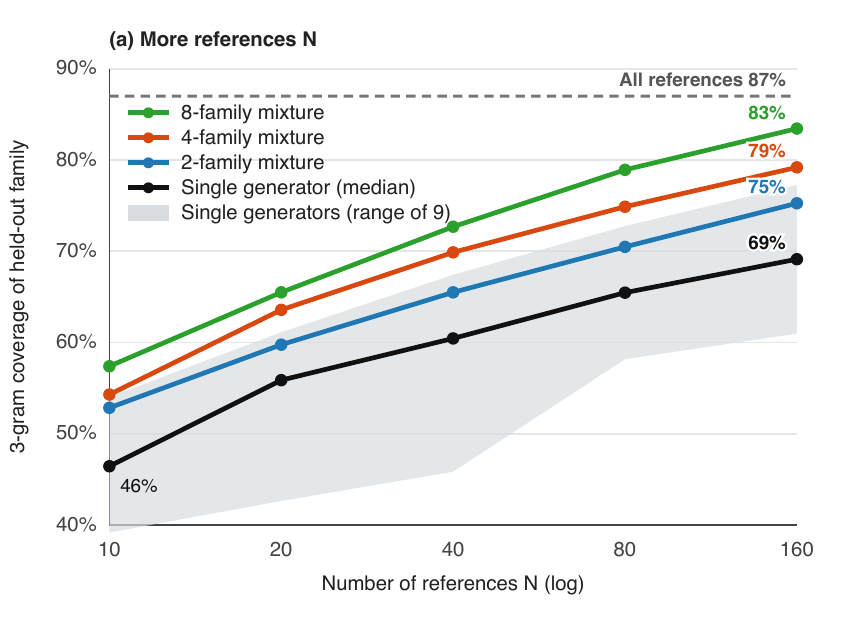}\hfill
  \includegraphics[height=6.2cm]{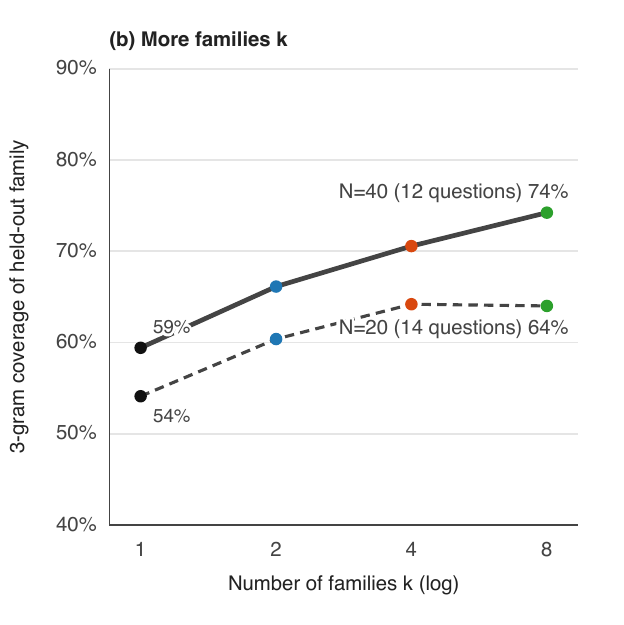}
  \caption{Held-out-family 3-gram coverage by reference count $N$ (a) and family count $k$ (b). The gray band spans the available single generators (only DeepSeek and Gemini at $N = 160$); panel (b) uses common questions across $k$ at each $N$.}
  \label{fig:coverage}
\end{figure*}

\section{Obtaining Valid and Varied References}
\label{sec:reference-analysis}

\subsection{Valid-Reference Yield}
\label{sec:yield}

\paragraph{Setup.}
We estimate valid-reference yield for base and post-trained variants of Gemma-4-31B and Qwen3-30B-A3B on the 15 Khmer and 15 Japanese analysis questions.
Each variant generates 100 candidates per question with eight-shot prompting, rather than the four-shot prompting of the main pool; base variants receive completion prompts and post-trained variants chat prompts, with examples and selection rules matched within each language (Appendix~\ref{app:rules}).
Because keyword policies were selected on post-trained candidates, the selection rules may favor post-trained acceptance (Appendix~\ref{app:selection}).
To estimate validity, one author manually checks 20--50 retained references per condition (Appendix~\ref{app:human}).

\paragraph{Results.}
Table~\ref{tab:yield} shows higher point estimates of acceptance, validity, and yield for post-trained than for base variants in both languages, although the two Japanese Qwen validity estimates differ by one of 20 audited references.
Yield is lower in Khmer than in Japanese for every model and variant (0--30\% versus 29--68\%); because the question sets are not parallel, the gap may reflect language, regional content, or selection rules.
A higher yield than the base variant does not ensure enough valid references: post-trained Qwen reaches only 2\% yield in Khmer, so about 2,000 candidates per question would be needed for 40 valid references.
Generator selection therefore requires checking absolute yield in the target language.

\begin{table}[t]
  \centering
  \small
  \setlength{\tabcolsep}{3pt}
  \begin{tabular}{@{}lrrrrr@{}}
    \toprule
    & & \multicolumn{2}{c}{\textbf{Coverage}} & \multicolumn{2}{c}{\textbf{Distinct-3}} \\
    \cmidrule(lr){3-4} \cmidrule(lr){5-6}
    \textbf{Generator} & \textbf{$N$} & \textbf{Base} & \textbf{Post} & \textbf{Base} & \textbf{Post} \\
    \midrule
    \multicolumn{6}{@{}l}{\textbf{Japanese}} \\
    Gemma-4-31B   & 39 & 76\% & 48\% & 31\% & 14\% \\
    Qwen3-30B-A3B & 36 & 71\% & 60\% & 37\% & 24\% \\
    \midrule
    \multicolumn{6}{@{}l}{\textbf{Khmer}} \\
    Gemma-4-31B   & 20 & 58\% & 48\% & 54\% & 20\% \\
    \bottomrule
  \end{tabular}
  \caption{Coverage and distinct-3 at matched reference counts. Japanese results are the main comparison; Khmer is supplementary. $N$ is the mean count across included questions; Post denotes post-trained.}
  \label{tab:coverage}
\end{table}

\subsection{Coverage by Model Variant}
\label{sec:coverage}

\paragraph{Setup.}
Using the retained references from Section~\ref{sec:yield}, we compare coverage and distinct-3 between base and post-trained variants at matched reference counts.
For each question, both variants contribute the smaller of the two retained counts, capped at 40 for Japanese and 20 for Khmer and sampled with 100 random seeds; questions with fewer than ten retained references from either variant are excluded.
Japanese provides the main comparison; in Khmer, Qwen is omitted because of its low acceptance and validity, leaving a supplementary Gemma comparison on nine questions.
The Japanese comparison answers come from base models and may favor base-variant references, whereas the Khmer comparison answers come from post-trained models outside the Gemma family.
Both measures count invalid retained references and are sensitive to total text length, which matched reference counts do not control.

\paragraph{Results.}
Table~\ref{tab:coverage} shows lower coverage and distinct-3 for post-trained than for base variants of both models in Japanese.
The Khmer Gemma comparison follows the same pattern, even though the Khmer comparison answers come from post-trained models.
Post-trained variants thus combine higher estimated yield (Table~\ref{tab:yield}) with lower 3-gram coverage of the comparison answers, so generator selection should assess yield and coverage separately.

\begin{figure*}[!t]
  \centering
  \includegraphics[width=0.85\textwidth]{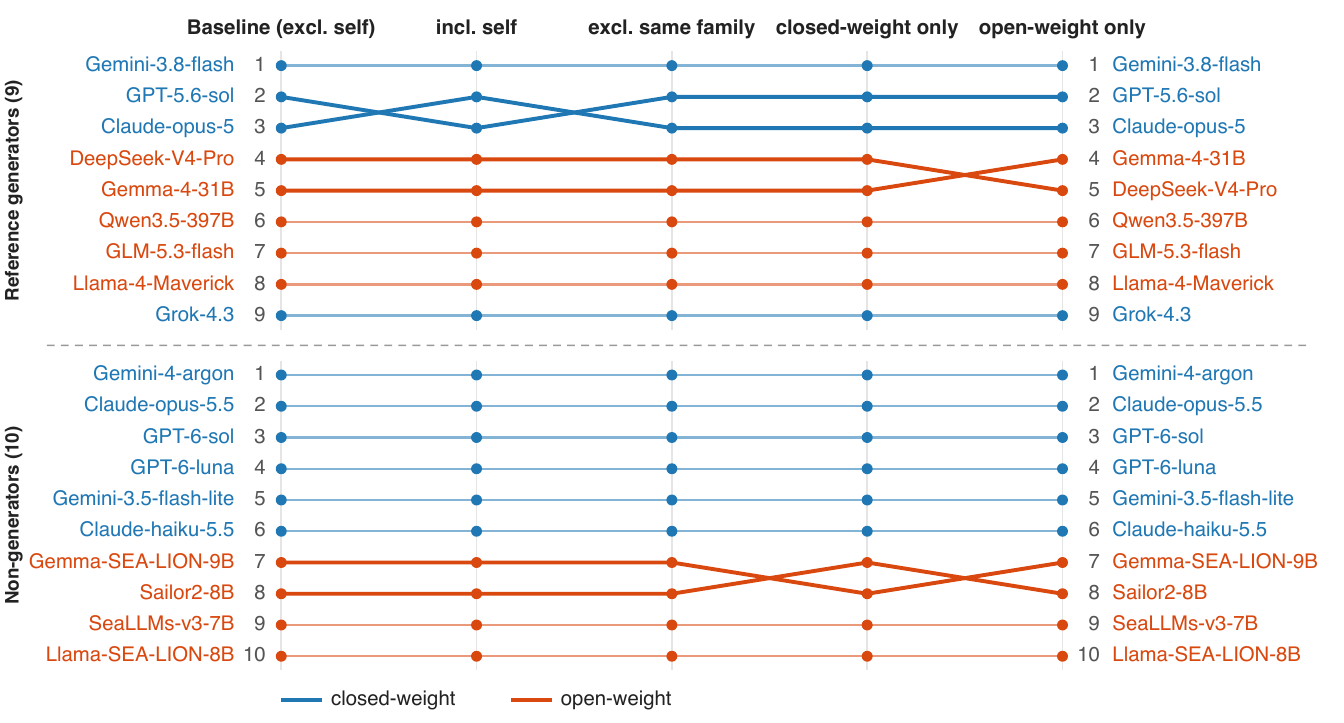}
  \caption{Rankings under alternative reference sets for the reference generators (top) and non-generators (bottom). Ranks are computed within each group; the leftmost condition is self-excluded scoring.}
  \label{fig:rank}
\end{figure*}

\subsection{Reference Count and Family Mixtures}
\label{sec:mixing}

\paragraph{Setup.}
Because post-trained variants had lower 3-gram coverage (Section~\ref{sec:coverage}) and all nine main-pool generators are post-trained, we test whether spreading a fixed number of references across families broadens coverage.
On the 15 Khmer analysis questions, each generator family in turn supplies the comparison answers, and $N$ references are drawn from the remaining families, either from one generator or as evenly as possible across $k$ families.
Results are aggregated over held-out families and 100 random seeds.
Available generators and questions differ across $N$, whereas comparisons across $k$ hold questions fixed at each $N$.

\paragraph{Results.}
Figure~\ref{fig:coverage}(a) shows coverage rising with $N$ in every configuration, and every family mixture lies above the single-generator median at each $N$.
With questions fixed (Figure~\ref{fig:coverage}(b)), coverage at $N = 40$ rises from 59\% with one family to 74\% with eight, whereas coverage at $N = 20$ plateaus from four to eight families.
Spreading a fixed number of post-trained references across families thus broadens coverage, although the gain from additional families depends on $N$.

\section{Rankings and Human Agreement}
\label{sec:validation}

\subsection{Reference Allocation and Rankings}
\label{sec:allocation}

\begin{table}[t]
  \centering
  \small
  \setlength{\tabcolsep}{4pt}
  \begin{tabular}{rrrrrr}
    \toprule
    \textbf{$N$} & \textbf{$k$} & \begin{tabular}[b]{@{}r@{}}\textbf{Per}\\\textbf{family}\end{tabular} & \textbf{Coverage} & \textbf{Rank corr.} & \textbf{Questions} \\
    \midrule
    40  & 1 & 40       & 64--76\% & 0.96--0.99 & 36--50 \\
        & 2 & 20       & 80--84\% & 0.98--1.00 & 32--47 \\
        & 4 & 10       & 87--89\% & 0.95--1.00 & 30--40 \\
    80  & 4 & 20       & 89--91\% & 0.97--1.00 & 30--39 \\
    160 & 4 & 40       & 93--94\% & 0.97--1.00 & 31--38 \\
    \midrule
    All & 9 & $\approx 40$ & 94\% & 1.00       & 50 \\
    \bottomrule
  \end{tabular}
  \caption{Coverage of eligible development gold answers and rank correlation with the full-pool ranking, as ranges over selected family combinations. Questions counts the questions used for the 19-model rankings; for All, $\approx 40$ is the mean per family per question. References are pooled with self-references retained.}
  \label{tab:allocation}
\end{table}

\paragraph{Setup.}
We test whether smaller reference sets, with families chosen to maximize coverage of the development gold answers, reproduce the full-pool ranking.
Each configuration distributes $N$ references per question equally across $k$ such families.
A question is eligible only if every selected family has at least $N/k$ references for it, and sampled-set and full-pool rankings use the same eligible questions.
Unlike the leaderboard, sampled-set and full-pool scores pool references into one group and retain self-references (Appendix~\ref{app:scoring}).
Coverage values use different eligible question subsets, so coverage differences between configurations do not isolate family-count effects.
A separate analysis varies $N$ with four families and 39 questions fixed, using all references from those families as the baseline (Appendix~\ref{app:allocation}).

\paragraph{Results.}
Table~\ref{tab:allocation} shows Spearman rank correlations of 0.95--1.00 with the full-pool ranking in every configuration, including single-family sets at $N = 40$.
With families and questions fixed, correlations with the four-family baseline round to 1.000 at $N = 40$, 80, and 160.
Forty references per question are therefore enough to reproduce the full-pool ranking of the 19 tested models.

\begin{figure*}[!t]
  \centering
  \includegraphics[width=0.9\textwidth]{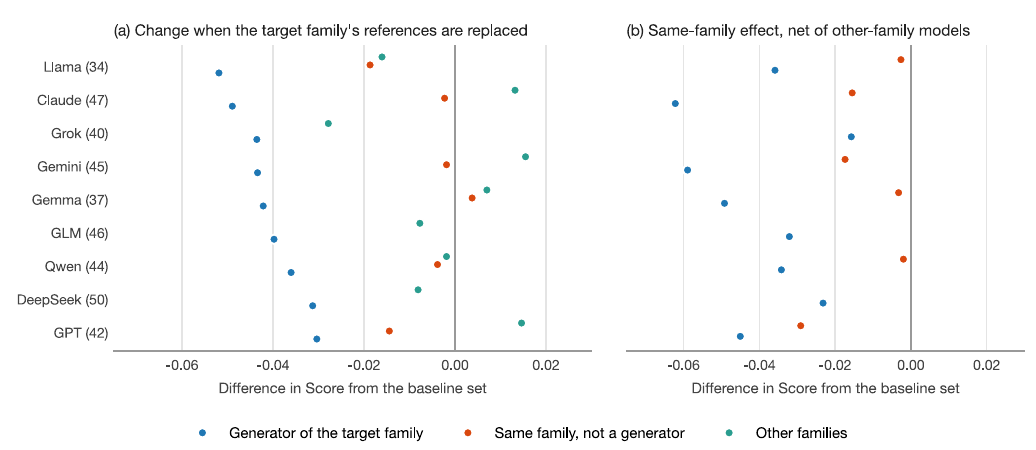}
  \caption{Score changes from replacing one family's references in a four-family set ($N = 40$): (a) by model role; (b) target-family models, net of other-family models. Dots are means over questions (counts in parentheses).}
  \label{fig:swap}
\end{figure*}

\begin{figure*}[!t]
  \centering
  \begin{subfigure}[b]{0.32\textwidth}
    \includegraphics[width=\linewidth]{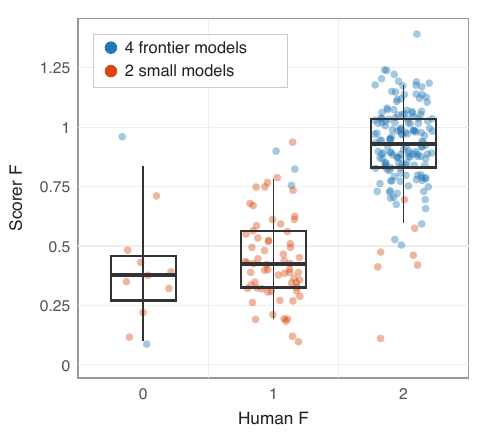}
    \caption{}
    \label{fig:human-f}
  \end{subfigure}\hfill
  \begin{subfigure}[b]{0.32\textwidth}
    \includegraphics[width=\linewidth]{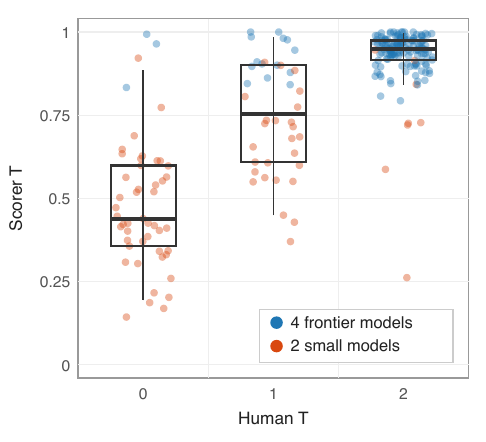}
    \caption{}
    \label{fig:human-t}
  \end{subfigure}\hfill
  \begin{subfigure}[b]{0.32\textwidth}
    \includegraphics[width=\linewidth]{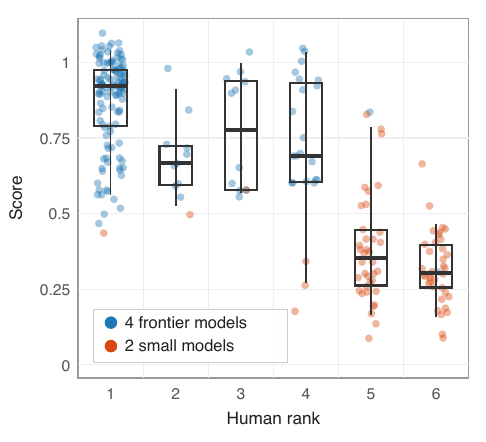}
    \caption{}
    \label{fig:human-rank}
  \end{subfigure}
  \caption{F and T by human rating (a, b) and Score by human rank (c; 1 = best). Frontier: Gemini-4-argon, GPT-6-sol, Claude-opus-5.5, GPT-6-luna; small: Gemma-SEA-LION-9B, Sailor2-8B. Whiskers: 5th--95th percentiles.}
  \label{fig:human}
\end{figure*}

\subsection{Dependence on Reference Origin}
\label{sec:stability}

\paragraph{Setup.}
We test whether scores depend on which models generated the references.
We compare the self-excluded leaderboard baseline with self-inclusion, exclusion of the evaluated model's family, and reference pools restricted to open-weight or closed-weight generators.
In a separate four-family pool with ten references per family, we replace a target family with an outside family.
The control replaces a non-target family with the same outside family.
Replacement comparisons retain self-references and use one pooled reference group (Appendix~\ref{app:replacement}).

\paragraph{Results.}
Figure~\ref{fig:rank} shows that rankings within the generator and non-generator groups change little across the alternative reference sets.
In the combined 19-model ranking, self-inclusion raises generator scores by up to 0.04 and changes eight ranks, mostly by moving generators above non-generators (Appendix Table~\ref{tab:leaderboard}).
The self-reference gain supports scoring each model without that model's own references, as the leaderboard does (Section~\ref{sec:scoring}).

Figure~\ref{fig:swap} shows that replacing a generator's family lowers that generator's score by about 0.03--0.05, whereas control replacements change scores by less than 0.01.
Same-family non-generators also decline relative to other-family models, most strongly for GPT, Gemini, and Claude.
Reference origin thus affects scores at the family level as well as the model level.

\subsection{Agreement with Human Ratings}
\label{sec:human}

\paragraph{Setup.}
One native Khmer-speaking author rates one uniformly sampled response per model--question pair for six non-generators on 40 test questions (240 responses).
The six models comprise four closed-weight frontier models and two small open-weight models (8B and 9B parameters); all families occur in the reference pool.
With model identities and benchmark scores hidden, the rater assigns 0--2 scores for fluency, correctness, and whether the response answers the question and ranks responses within questions.
Pairwise agreement is the fraction of within-question response pairs that Score orders as the human ranking does, averaged over questions; human-tied pairs are excluded and Score ties count 0.5.

\paragraph{Results.}
Figure~\ref{fig:human}(a, b) shows median F and T scores increasing with the corresponding human ratings.
Figure~\ref{fig:human}(c) shows lower Score for responses at human ranks 5 and 6, mostly from the two small models, than at ranks 1--4.
Across the six models, the human ranking correlates at $\rho = 0.94$ with the ranking by Score on these responses and at $\rho = 0.83$ with the leaderboard ranking (Appendix Table~\ref{tab:leaderboard}).
Within-question pairwise agreement is 87.9\% overall, 97.4\% for frontier-versus-small-model pairs, and 59.8\% among frontier responses (chance: 50\%).
Agreement with the native speaker thus comes mainly from separating frontier from small models.

\section{Conclusion}

We introduced Khmer PFGen-Bench, a judge-free benchmark for open-ended generation in Khmer, whose model rankings remain highly correlated across reference sets and agree with a native speaker at the model level.
Building the benchmark showed that valid and varied references cannot be taken for granted in a low-resource language: generators need to be checked for both valid-reference yield and coverage in the target language and drawn from several families, and each model should be scored without its own references.
These practices, together with validation by independent native speakers, offer a path to judge-free evaluation in other low-resource languages.

\section*{Limitations}

The study covers 50 questions and a limited set of generators, prompts, and selection rules.
The variant comparisons cover only two model pairs, with completion prompts for base variants and chat prompts for post-trained variants, so the comparisons describe the available variants as prompted rather than an effect of post-training; the Japanese and Khmer question sets are also non-parallel.
Validity audits cover at most 50 references per condition, all from the variant comparisons rather than the main reference pool, leaving substantial uncertainty in validity and yield estimates.
Development gold answers both select and assess the allocation configurations, so development coverage is not independent validation.
Coverage records 3-gram presence, whereas scoring also uses reference frequencies, and whether broader coverage improves agreement with human judgments remains untested.
Models from families outside the reference pool are not evaluated, and the scorer is not compared with LLM judges on the same human ratings.

Human validation uses one native Khmer-speaking author who also wrote the questions, gold answers, and keyword rules, so inter-rater agreement cannot be measured, and author familiarity may influence the ratings despite blinding.
The six rated models are all non-generators, so the human ratings do not test comparisons involving generators, where self-inclusion changes ranks.

\section*{Ethics Statement}

The questions cover general-education topics from the Cambodian school curriculum and contain no personal data.
All human ratings were produced by the native Khmer-speaking author; no external annotators were recruited.
References in the main pool passed rule-based selection but were not individually verified, so they may contain factual errors.
The benchmark is intended for model comparison, not as an authoritative source of answers.

\bibliography{references}

@incollection{chea-etal-2007-word,
  title     = {Word Bigram Vs Orthographic Syllable Bigram in {Khmer} Word Segmentation},
  author    = {Chea, Sok Huor and Top, Rithy and Ros, Pich Hemy and Vann, Navy and Chin, Chanthirith and Chhoeun, Tola},
  booktitle = {{PAN} Localization Working Papers 2004--2007},
  publisher = {PAN Localization Project},
  year      = {2007},
  url       = {https://www.researchgate.net/publication/228972610_Word_Bigram_Vs_Orthographic_Syllable_Bigram_in_Khmer_Word_Segmentation}
}

@inproceedings{hada-etal-2024-large,
  title     = {Are Large Language Model-based Evaluators the Solution to Scaling Up Multilingual Evaluation?},
  author    = {Hada, Rishav and Gumma, Varun and de Wynter, Adrian and Diddee, Harshita and Ahmed, Mohamed and Choudhury, Monojit and Bali, Kalika and Sitaram, Sunayana},
  booktitle = {Findings of the Association for Computational Linguistics: EACL 2024},
  year      = {2024}
}

@techreport{hosken-2025-khmer,
  title       = {{Khmer} Encoding Structure},
  author      = {Hosken, Martin},
  institution = {Unicode Consortium},
  type        = {Unicode Technical Note},
  number      = {61},
  year        = {2025},
  note        = {Version 2},
  url         = {https://www.unicode.org/notes/tn61/tn61-2.html}
}

@article{imajo-etal-2025-judge,
  title   = {A Judge-free {LLM} Open-ended Generation Benchmark Based on the Distributional Hypothesis},
  author  = {Imajo, Kentaro and Hirano, Masanori and Suzuki, Shuji and Mikami, Hiroaki},
  journal = {arXiv preprint arXiv:2502.09316},
  year    = {2025},
  url     = {https://arxiv.org/abs/2502.09316}
}

@inproceedings{kew-etal-2024-turning,
  title     = {Turning {E}nglish-centric {LLM}s Into Polyglots: How Much Multilinguality Is Needed?},
  author    = {Kew, Tannon and Schottmann, Florian and Sennrich, Rico},
  booktitle = {Findings of the Association for Computational Linguistics: EMNLP 2024},
  year      = {2024}
}

@inproceedings{kirk-etal-2024-understanding,
  title     = {Understanding the Effects of {RLHF} on {LLM} Generalisation and Diversity},
  author    = {Kirk, Robert and Mediratta, Ishita and Nalmpantis, Christoforos and Luketina, Jelena and Hambro, Eric and Grefenstette, Edward and Raileanu, Roberta},
  booktitle = {Proceedings of the Twelfth International Conference on Learning Representations (ICLR 2024)},
  year      = {2024}
}

@inproceedings{li-etal-2016-diversity,
  title     = {A Diversity-Promoting Objective Function for Neural Conversation Models},
  author    = {Li, Jiwei and Galley, Michel and Brockett, Chris and Gao, Jianfeng and Dolan, Bill},
  booktitle = {Proceedings of the 2016 Conference of the North American Chapter of the Association for Computational Linguistics: Human Language Technologies},
  year      = {2016},
  pages     = {110--119},
  doi       = {10.18653/v1/N16-1014},
  url       = {https://aclanthology.org/N16-1014/}
}

@inproceedings{shaham-etal-2024-multilingual,
  title     = {Multilingual Instruction Tuning With Just a Pinch of Multilinguality},
  author    = {Shaham, Uri and Herzig, Jonathan and Aharoni, Roee and Szpektor, Idan and Tsarfaty, Reut and Eyal, Matan},
  booktitle = {Findings of the Association for Computational Linguistics: ACL 2024},
  year      = {2024}
}

@article{son-etal-2024-mmeval,
  title   = {{MM-Eval}: A Multilingual Meta-Evaluation Benchmark for {LLM}-as-a-Judge and Reward Models},
  author  = {Son, Guijin and Yoon, Dongkeun and Suk, Juyoung and others},
  journal = {arXiv preprint arXiv:2410.17578},
  year    = {2024}
}

@inproceedings{han-etal-2026-deepsea,
  title     = {{D}eep{SEA}: A Multilingual Evaluation Benchmark for {S}outheast {A}sia},
  author    = {Han, Jiuzhou and Mishra, Abhishek Bijay and Cohn, Trevor and Mittal, Shubham},
  booktitle = {Proceedings of the 2026 Conference on Empirical Methods in Natural Language Processing},
  pages     = {1--17},
  year      = {2026}
}

@inproceedings{nguyen-etal-2024-seallms,
  title     = {{S}ea{LLM}s -- Large Language Models for {S}outheast {A}sia},
  author    = {Nguyen, Xuan-Phi and Zhang, Wenxuan and Li, Xin and Aljunied, Mahani and others},
  booktitle = {Proceedings of the 62nd Annual Meeting of the Association for Computational Linguistics (Volume 3: System Demonstrations)},
  pages     = {294--304},
  year      = {2024}
}

@misc{hupkes-bogoychev-2025-multiloko,
  title         = {{MultiLoKo}: a multilingual local knowledge benchmark for {LLM}s spanning 31 languages},
  author        = {Hupkes, Dieuwke and Bogoychev, Nikolay},
  year          = {2025},
  eprint        = {2504.10356},
  archivePrefix = {arXiv},
  primaryClass  = {cs.CL}
}

@inproceedings{liu-etal-2025-seaexam,
  title     = {{S}ea{E}xam and {S}ea{B}ench: Benchmarking {LLM}s with Local Multilingual Questions in {S}outheast {A}sia},
  author    = {Liu, Chaoqun and Zhang, Wenxuan and Ying, Jiahao and Aljunied, Mahani and Luu, Anh Tuan and Bing, Lidong},
  booktitle = {Findings of the Association for Computational Linguistics: NAACL 2025},
  pages     = {6134--6151},
  year      = {2025}
}

@inproceedings{susanto-etal-2025-seahelm,
  title     = {{SEA-HELM}: {S}outheast {A}sian Holistic Evaluation of Language Models},
  author    = {Susanto, Yosephine and Hulagadri, Adithya Venkatadri and Montalan, Jann Railey and others},
  booktitle = {Findings of the Association for Computational Linguistics: ACL 2025},
  pages     = {12308--12336},
  year      = {2025}
}

@inproceedings{tang-etal-2024-metrics,
  title     = {Not All Metrics Are Guilty: Improving {NLG} Evaluation by Diversifying References},
  author    = {Tang, Tianyi and Lu, Hongyuan and Jiang, Yuchen Eleanor and Huang, Haoyang and Zhang, Dongdong and Zhao, Wayne Xin and Kocmi, Tom and Wei, Furu},
  booktitle = {Proceedings of the 2024 Conference of the North American Chapter of the Association for Computational Linguistics: Human Language Technologies (Volume 1: Long Papers)},
  year      = {2024}
}

@inproceedings{zeng-etal-2024-towards,
  title     = {Towards Multiple References Era -- Addressing Data Leakage and Limited Reference Diversity in Machine Translation Evaluation},
  author    = {Zeng, Xianfeng and Liu, Yijin and Meng, Fandong and Zhou, Jie},
  booktitle = {Findings of the Association for Computational Linguistics: ACL 2024},
  pages     = {11939--11951},
  year      = {2024}
}

@inproceedings{zheng-etal-2023-judging,
  title     = {Judging {LLM}-as-a-Judge with {MT-Bench} and {C}hatbot {A}rena},
  author    = {Zheng, Lianmin and Chiang, Wei-Lin and Sheng, Ying and others},
  booktitle = {Advances in Neural Information Processing Systems 36 (NeurIPS 2023), Datasets and Benchmarks Track},
  year      = {2023}
}

@misc{moeys-2015-curriculum,
  title        = {{Curriculum Framework of General Education and Technical Education}},
  author       = {{Ministry of Education, Youth and Sport}},
  howpublished = {Department of Curriculum Development, Phnom Penh},
  year         = {2015},
  url          = {https://moeys.gov.kh/en/policies-and-strategic-plans/document-5416}
}

\clearpage
\appendix

\counterwithin*{table}{section}
\renewcommand{\thetable}{\thesection\arabic{table}}
\renewcommand{\theHtable}{\thesection.\arabic{table}}
\counterwithin*{figure}{section}
\renewcommand{\thefigure}{\thesection\arabic{figure}}
\renewcommand{\theHfigure}{\thesection.\arabic{figure}}

\makeatletter
\setlength{\@fptop}{0pt}
\setlength{\@fpsep}{16pt}
\setlength{\@fpbot}{0pt plus 1fil}
\setlength{\@dblfptop}{0pt}
\setlength{\@dblfpsep}{16pt}
\setlength{\@dblfpbot}{0pt plus 1fil}
\makeatother

\section{Reference Construction}
\label{app:rules}

\subsection{Prompts and Sampling}
\label{app:prompts}

Table~\ref{tab:prompts} shows the prompts for reference generation and model evaluation.
Reference generation uses four question--answer examples from other categories, fixed for each target question and shared across generators.
Generators sample with temperature 1.0 and top-$p$ 0.98, except Claude-opus-5, which uses API defaults; empty or truncated outputs are discarded.

\begin{table*}[t]
  \centering
  \small
  \begin{tabular}{@{}p{0.47\textwidth}@{\hspace{0.06\textwidth}}p{0.47\textwidth}@{}}
    \toprule
    \textbf{Reference generation: conversation turns} & \textbf{Model evaluation: one user message} \\
    \midrule
    \raggedright
    \textrm{User: <example question>}\newline
    \khmer{ឆ្លើយខ្លីៗ ២-៣ ប្រយោគ។}\newline
    \textit{[Answer briefly in 2--3 sentences.]}\newline
    \textrm{Assistant: <example gold answer>}\newline
    \textit{[Repeat for four examples.]}\newline\newline
    \textrm{User: <target question>}\newline
    \khmer{ឆ្លើយខ្លីៗ ២-៣ ប្រយោគ។}
    &
    \raggedright
    \khmer{ចូរឆ្លើយសំណួរក្នុងមួយបន្ទាត់ ដោយប្រើរបៀបសរសេរ និងប្រវែងប្រហាក់ប្រហែលឧទាហរណ៍។}\newline
    \textit{[Answer the question in one line, using a style and length similar to the examples.]}\newline\newline
    \khmer{\lat{\#\#} ឧទាហរណ៍} \textit{[Examples]}\newline
    \textrm{Q: <example question>}\newline
    \textrm{A: <example gold answer>}\newline
    \textit{[Repeat for eight examples.]}\newline\newline
    \khmer{\lat{\#\#} សំណួរ} \textit{[Question]}\newline
    \textrm{Q: <target question>} \tabularnewline
    \bottomrule
  \end{tabular}
  \caption{Prompt structures. Translations, repetition notes, role labels, and angle-bracketed placeholders are explanatory. Reference generation uses conversation turns, with the instruction in the same user turn as each question; evaluation uses one user message. The target answer is never supplied.}
  \label{tab:prompts}
\end{table*}

\paragraph{Base and post-trained variants.}
The variant comparisons (Section~\ref{sec:yield}) sample 100 candidates per question with eight examples.
Post-trained variants receive the examples as conversation turns (Table~\ref{tab:prompts}, left); base variants receive the same examples as ``Q:'' and ``A:'' pairs and complete an empty ``A:'' for the target question.

\subsection{Normalization and Selection Rules}
\label{app:selection}

Normalization applies Unicode NFC, removes invisible characters, whitespace, and punctuation, canonicalizes legacy Khmer sequences \citep{hosken-2025-khmer}, and unifies digits; evaluation responses are normalized in the same way before scoring.
A KCC is a base character with its following subscript consonants, dependent vowels, and combining marks.

Four rules are applied in sequence (Table~\ref{tab:funnel}):
(1) format, rejecting Markdown markup, runs of seven or more Latin-script words, and unrelated scripts such as Thai or Cyrillic;
(2) duplication, keeping one copy of each identical normalized answer per question;
(3) length, keeping 40--201 KCCs for Khmer or 30--150 characters for Japanese, where the upper bound is the zero point of the scorer's length discount; and
(4) keywords.
The keyword rules contain 216 elements derived from the gold answers, each with alternative phrasings.
Each question's acceptance policy requires either all required elements (31 questions) or at least one element (19 questions), choosing the stricter policy that still keeps at least 50 main-pool references from each of the open-weight and closed-weight groups.
The Japanese comparisons use the original benchmark's keywords, with policies chosen in the same way on post-trained Gemma candidates.

\begin{table}[t]
  \centering
  \begin{tabular}{lr}
    \toprule
    \textbf{Stage} & \textbf{Remaining} \\
    \midrule
    Candidates  & 29,321 \\
    Format      & 27,986 \\
    Duplication & 26,212 \\
    Length      & 25,514 \\
    Keywords    & 18,274 \\
    \bottomrule
  \end{tabular}
  \caption{Candidates retained after each filtering stage.}
  \label{tab:funnel}
\end{table}

\subsection{Reference Counts}

Table~\ref{tab:per-generator} gives counts by generator.
Of the 450 generator--question pairs, 376 contain at least 40 references, and every question has at least four families with 40 or more.

\begin{table}[!ht]
  \centering
  \small
  \setlength{\tabcolsep}{3pt}
  \begin{tabular}{@{}lrrr@{}}
    \toprule
    \textbf{Generator} & \textbf{Candidates} & \textbf{Retained} & \textbf{Acceptance} \\
    \midrule
    DeepSeek-V4-Pro   & 3,060  & 2,371  & 77\% \\
    Llama-4-Maverick  & 3,650  & 1,691  & 46\% \\
    Gemma-4-31B    & 2,910  & 1,693  & 58\% \\
    Qwen3.5-397B & 3,060  & 2,074  & 68\% \\
    GLM-5.3-flash     & 3,341  & 2,150  & 64\% \\
    \midrule
    GPT-5.6-sol       & 3,580  & 2,014  & 56\% \\
    Grok-4.3          & 4,060  & 1,784  & 44\% \\
    Claude-opus-5     & 2,800  & 2,284  & 82\% \\
    Gemini-3.8-flash  & 2,860  & 2,213  & 77\% \\
    \midrule
    Total             & 29,321 & 18,274 & 62\% \\
    \bottomrule
  \end{tabular}
  \caption{Candidate counts, retained references, and acceptance rates for the final reference pool. The first five generators are open-weight and the remaining four closed-weight. Acceptance denotes rule-based retention, not audited validity.}
  \label{tab:per-generator}
\end{table}

\section{Evaluation Data and Protocol}
\label{app:setup}

\begin{table*}[t]
  \centering
  \begin{tabular}{lllr}
    \toprule
    \textbf{Model} & \textbf{Role} & \textbf{Family} & \textbf{Failures} \\
    \midrule
    DeepSeek-V4-Pro                        & generator     & DeepSeek &  0 \\
    Llama-4-Maverick-17B-128E-Instruct-FP8 & generator     & Llama    &  0 \\
    Gemma-4-31B-it                         & generator     & Gemma    &  0 \\
    Qwen3.5-397B-A17B                      & generator     & Qwen     &  0 \\
    GLM-5.3-flash                          & generator     & GLM      & 13 \\
    GPT-5.6-sol                            & generator     & GPT      &  0 \\
    Grok-4.3                               & generator     & Grok     &  0 \\
    Claude-opus-5                          & generator     & Claude   &  0 \\
    Gemini-3.8-flash                       & generator     & Gemini   &  0 \\
    \midrule
    Gemini-4-argon                         & non-generator        & Gemini   &  0 \\
    GPT-6-sol                              & non-generator        & GPT      &  0 \\
    Claude-opus-5.5                        & non-generator        & Claude   &  0 \\
    GPT-6-luna                             & non-generator        & GPT      &  0 \\
    Gemini-3.5-flash-lite                  & non-generator        & Gemini   &  0 \\
    Claude-haiku-5.5                       & non-generator        & Claude   &  0 \\
    Gemma-SEA-LION-v3-9B-IT                & non-generator        & Gemma    &  0 \\
    Sailor2-8B-Chat                        & non-generator        & Qwen     &  0 \\
    SeaLLMs-v3-7B-Chat                     & non-generator        & Qwen     &  0 \\
    Llama-SEA-LION-v3-8B-IT                & non-generator        & Llama    & 12 \\
    \bottomrule
  \end{tabular}
  \caption{Evaluated models, roles, and families. Failures are responses scored zero, out of 500 per model (50 questions, ten trials). Short model names are used elsewhere in the paper.}
  \label{tab:models}
\end{table*}

\subsection{Questions and Splits}

The 50 questions were written directly in Khmer: 27 on regional topics (Khmer language, history, geography, civics, arts, and culture) and 23 on general subjects (mathematics, physics, chemistry, biology, earth science, health, and information technology), as listed in Tables~\ref{tab:questions-a} and~\ref{tab:questions-b}.
A question's own gold answer never appears in the prompts for that question.
The 15 Khmer analysis questions are sampled across categories, eight regional and seven general; the 15 Japanese questions come from the original benchmark.
The development--test split was fixed before human rating, with five of the 15 Khmer analysis questions in development and ten in test.

\begin{table*}[!tp]
  \centering
  \small
  \setlength{\tabcolsep}{4pt}
  \begin{tabular}{@{}>{\raggedright\arraybackslash}p{0.105\textwidth}l>{\raggedright\arraybackslash}p{0.455\textwidth}>{\raggedright\arraybackslash}p{0.33\textwidth}@{}}
    \toprule
    \textbf{Category} & \textbf{ID} & \textbf{Question} & \textbf{English translation} \\
    \midrule
    Khmer language & Q01 & \khmer{ចូររៀបរាប់អំពីពាក្យក្លាយក្នុងភាសាខ្មែរ។} & Describe derived words in Khmer. \\
     & Q02 & \khmer{ចូររៀបរាប់កាព្យមាត្រនិងចំណាប់ចួននៃកំណាព្យបទភុជង្គលីលា។} & Describe the meter and rhyme scheme of \textit{Phujong Lilea} verse. \\
     & Q03 & \khmer{តើរឿងព្រះអាទិត្យថ្មីរះលើផែនដីចាស់និយាយអំពីអ្វី?} & What is the novel \textit{A New Sun Rises over the Old Land} about? \\
     & Q04 & \khmer{ចូរបង្ហាញភាពខុសគ្នារវាងសុភាសិតនិងពាក្យស្លោក។} & Explain the difference between proverbs and slogans. \\
     & Q05 & \khmer{ចូររៀបរាប់អំពីលក្ខណៈសម្គាល់របស់អក្សរសិល្ប៍ខ្មែរបែបពុទ្ធនិយម។} & Describe the distinctive features of Buddhist Khmer literature. \\
    \midrule
    Mathematics & Q06 & \khmer{តើគេអាចសម្គាល់ថាត្រីកោណសាមញ្ញពីរប៉ុនគ្នា តាមលក្ខខណ្ឌអ្វីខ្លះ?} & By which criteria can two general triangles be identified as congruent? \\
     & Q07 & \khmer{ដូចម្តេចដែលហៅថាចំនួនបឋម?} & What is a prime number? \\
     & Q08 & \khmer{ក្នុងរង្វាស់ប្រវែងបុរាណខ្មែរ តើ១ហត្ថស្មើប៉ុន្មានម៉ែត្រ?} & In traditional Khmer units of length, how many meters is one \textit{hat} (cubit)? \\
     & Q09 & \khmer{តើប្រូបាបនៃព្រឹត្តិការណ៍មួយ និងព្រឹត្តិការណ៍បំពេញរបស់វាមានទំនាក់ទំនងគ្នាយ៉ាងដូចម្តេច?} & How is the probability of an event related to that of its complementary event? \\
     & Q10 & \khmer{តើពហុកោណនិយ័តមានលក្ខណៈដូចម្តេចខ្លះ?} & What are the properties of a regular polygon? \\
    \midrule
    Physics & Q11 & \khmer{តើកម្តៅនិងសីតុណ្ហភាពខុសគ្នាដូចម្តេច?} & How do heat and temperature differ? \\
     & Q12 & \khmer{តើចរន្តជាប់និងចរន្តឆ្លាស់ខុសគ្នាដូចម្តេច?} & How do direct current and alternating current differ? \\
     & Q13 & \khmer{ដូចម្តេចហៅថាទន្លាក់សេរី?} & What is free fall? \\
    \midrule
    Chemistry & Q14 & \khmer{តើអ៊ីសូតូបជាអ្វី?} & What is an isotope? \\
     & Q15 & \khmer{តើតម្លៃ \lat{pH} បង្ហាញពីលក្ខណៈអ្វីរបស់សូលុយស្យុង?} & What property of a solution does the pH value indicate? \\
     & Q16 & \khmer{តើច្បាប់រក្សាម៉ាសចែងថាដូចម្តេច?} & What does the law of conservation of mass state? \\
    \midrule
    Biology & Q17 & \khmer{អ្វីទៅជាដំណើររស្មីសំយោគ?} & What is photosynthesis? \\
     & Q18 & \khmer{តើវីរុសនិងបាក់តេរីខុសគ្នាដូចម្តេច?} & How do viruses and bacteria differ? \\
     & Q19 & \khmer{តើអាំងស៊ុយលីនមានតួនាទីអ្វីក្នុងការគ្រប់គ្រងជាតិស្ករក្នុងឈាម?} & What role does insulin play in regulating blood sugar? \\
    \midrule
    Earth science & Q20 & \khmer{តើសូរ្យគ្រាសនិងចន្ទគ្រាសខុសគ្នាដូចម្តេច?} & How do solar and lunar eclipses differ? \\
     & Q21 & \khmer{ចូរបង្ហាញភាពខុសគ្នារវាងរលក\lat{P}និងរលក\lat{S}។} & Explain the difference between P waves and S waves. \\
     & Q22 & \khmer{តើធាតុអាកាសនិងអាកាសធាតុខុសគ្នាដូចម្តេច?} & How do weather and climate differ? \\
    \midrule
    History & Q23 & \khmer{តើមានការលើកឡើងដូចម្តេចខ្លះអំពីដើមកំណើតរដ្ឋខ្មែរដំបូង?} & What accounts have been proposed for the origin of the first Khmer state? \\
     & Q24 & \khmer{ចូររៀបរាប់អំពីស្នាព្រះហស្តរបស់ព្រះបាទជ័យវរ្ម័នទី៧ក្នុងប្រវត្តិសាស្ត្រខ្មែរ។} & Describe the achievements of King Jayavarman VII in Khmer history. \\
     & Q25 & \khmer{តើចលនាតស៊ូដើម្បីឯករាជ្យជាតិក្នុងប្រទេសកម្ពុជានាសម័យអាណានិគមនិយមបារាំងចែកចេញជាប៉ុន្មានផ្នែក?} & Into how many parts was the movement for national independence in Cambodia divided during the French colonial period? \\
    \bottomrule
  \end{tabular}
  \caption{Questions Q01--Q25 in the original Khmer, with English translations for readers.}
  \label{tab:questions-a}
\end{table*}

\begin{table*}[!tp]
  \centering
  \small
  \setlength{\tabcolsep}{4pt}
  \begin{tabular}{@{}>{\raggedright\arraybackslash}p{0.105\textwidth}l>{\raggedright\arraybackslash}p{0.455\textwidth}>{\raggedright\arraybackslash}p{0.33\textwidth}@{}}
    \toprule
    \textbf{Category} & \textbf{ID} & \textbf{Question} & \textbf{English translation} \\
    \midrule
    History & Q26 & \khmer{ដូចម្តេចដែលហៅថាសមាជជាតិ?} & What is the National Congress? \\
     & Q27 & \khmer{ក្នុងសម័យកម្ពុជាប្រជាធិបតេយ្យ តើពាក្យ }\guillemotleft\khmer{អង្គការ}\guillemotright\khmer{ មានន័យដូចម្តេច?} & In the Democratic Kampuchea period, what did the word \textit{Angkar} (organization) mean? \\
     & Q28 & \khmer{តើកិច្ចព្រមព្រៀងទីក្រុងប៉ារីសឆ្នាំ១៩៩១មានគោលបំណងសំខាន់ៗអ្វីខ្លះ?} & What were the main objectives of the 1991 Paris Peace Agreements? \\
    \midrule
    Geography & Q29 & \khmer{តើសណ្ឋានដីប្រទេសកម្ពុជាចែកចេញជាតំបន់អ្វីខ្លះ?} & Into which regions is the topography of Cambodia divided? \\
     & Q30 & \khmer{ហេតុអ្វីបានជាបឹងទន្លេសាបមានទំហំខុសគ្នារវាងរដូវប្រាំងនិងរដូវវស្សា?} & Why does the size of Tonle Sap Lake differ between the dry and rainy seasons? \\
     & Q31 & \khmer{តើទន្លេមេគង្គហូរឆ្លងកាត់រាជធានីខេត្តណាខ្លះនៃប្រទេសកម្ពុជា?} & Through which of Cambodia's capital and provinces does the Mekong River flow? \\
     & Q32 & \khmer{តើខេត្តណាខ្លះនៃប្រទេសកម្ពុជាដែលមានទីតាំងស្ថិតនៅជាប់សមុទ្រ?} & Which provinces of Cambodia are located on the coast? \\
     & Q33 & \khmer{ដូចម្តេចដែលហៅថានគរូបនីយកម្ម?} & What is urbanization? \\
     & Q34 & \khmer{តើខ្យល់មូសុងនិរតីនិងខ្យល់មូសុងឦសានមានឥទ្ធិពលយ៉ាងណាលើអាកាសធាតុកម្ពុជា?} & How do the southwest and northeast monsoons affect the climate of Cambodia? \\
    \midrule
    Civics & Q35 & \khmer{ចូរពន្យល់អត្ថន័យនៃព្រហ្មវិហារធម៌ទាំង៤។} & Explain the meaning of the four \textit{Brahmavih\=aras}. \\
     & Q36 & \khmer{តើភេទនិងយេនឌ័រខុសគ្នាយ៉ាងដូចម្តេច?} & How do sex and gender differ? \\
     & Q37 & \khmer{តើរដ្ឋសភាមានតួនាទីអ្វីខ្លះ?} & What are the roles of the National Assembly? \\
    \midrule
    Arts & Q38 & \khmer{តើគេបែងចែកការរាំវង់ជាប៉ុន្មានចង្វាក់ធំៗ?} & Into how many main rhythms is the \textit{romvong} circle dance divided? \\
     & Q39 & \khmer{តើឧបករណ៍ចាប៉ីដងវែងត្រូវបានប្រើប្រាស់ក្នុងទម្រង់សិល្បៈណាខ្លះ?} & In which art forms is the \textit{chapei dang veng} (long-necked lute) used? \\
     & Q40 & \khmer{តើគេចែកចម្លាក់បុរាណខ្មែរជាប៉ុន្មានប្រភេទ?} & Into how many types are ancient Khmer sculptures classified? \\
    \midrule
    Culture & Q41 & \khmer{ចូររៀបរាប់ពីអត្ថប្រយោជន៍នៃក្រមាខ្មែរ។} & Describe the uses of the Khmer \textit{krama} (scarf). \\
     & Q42 & \khmer{តើរបៀបសំពះខ្មែរមានប៉ុន្មានបែប?អ្វីខ្លះ?} & How many forms of the Khmer \textit{sampeah} greeting are there, and what are they? \\
     & Q43 & \khmer{តាមប្រពៃណីខ្មែរ តើពណ៌ប្រចាំថ្ងៃក្នុងមួយសប្តាហ៍មានអ្វីខ្លះ?} & In Khmer tradition, what color is associated with each day of the week? \\
     & Q44 & \khmer{តើថ្ងៃនីមួយៗក្នុងពិធីចូលឆ្នាំខ្មែរមានឈ្មោះអ្វីខ្លះ? ហើយគេធ្វើអ្វីខ្លះនៅថ្ងៃនីមួយៗ?} & What is each day of the Khmer New Year celebration called, and what do people do on each day? \\
    \midrule
    Health & Q45 & \khmer{តើត្រីមានសារធាតុចិញ្ចឹមអ្វីខ្លះ?} & What nutrients does fish contain? \\
     & Q46 & \khmer{ហេតុអ្វីបានជាភាគល្អិតនិចល \lat{PM2.5} មានគ្រោះថ្នាក់ជាង \lat{PM10}?} & Why is PM2.5 particulate matter more harmful than PM10? \\
     & Q47 & \khmer{តើសន្ទស្សន៍ម៉ាសរាងកាយ (\lat{BMI}) ជាអ្វី?} & What is the body mass index (BMI)? \\
    \midrule
    Information technology & Q48 & \khmer{ចូរបង្ហាញភាពខុសគ្នារវាងបណ្តាញ\lat{LAN}និង\lat{WAN}។} & Explain the difference between LAN and WAN networks. \\
     & Q49 & \khmer{តើ\lat{VR}និង\lat{AR}ខុសគ្នាដូចម្តេច?} & How do VR and AR differ? \\
     & Q50 & \khmer{តើសមាសភាគសំខាន់ៗទាំង៤នៃកុំព្យូទ័រលើតុមានអ្វីខ្លះ?} & What are the four main components of a desktop computer? \\
    \bottomrule
  \end{tabular}
  \caption{Questions Q26--Q50 in the original Khmer, with English translations for readers.}
  \label{tab:questions-b}
\end{table*}

\subsection{Response Generation}

Evaluation uses the single-message prompt in Table~\ref{tab:prompts}, with eight examples drawn from the other 49 questions for each question and trial.
Sampling uses temperature 1.0 and top-$p$ 0.98, except Claude-opus-5, which uses API defaults; the four 7--9B models run locally with Q4\_K\_M quantization.
Failed generations score zero and remain in the mean.

\subsection{Scoring Configuration}
\label{app:scoring}

Scores are computed with the released PFGen-Bench code \citep{imajo-etal-2025-judge}.
For Khmer, the length discount uses $\ell_1 = 134$ and $\ell_0 = 201$ KCCs, H checks the first 268 KCCs, and T fully credits a response unit covered by a 3-gram that occurs in at least $\tau = 2\%$ of the references (0.5\% in the original Japanese setting).
At $N = 40$, this threshold is below one reference, so T credits any 3-gram that occurs in a single reference.
The leaderboard averages F and T over the open-weight and closed-weight reference groups and excludes the evaluated model's own references; the allocation and family-replacement experiments pool references into one group and retain self-references.

\section{Leaderboard and Analysis Settings}
\label{app:tables}

\begin{table*}[t]
  \centering
  \begin{tabular}{rlrrrrrc}
    \toprule
    \textbf{Rank} & \textbf{Model} & \textbf{Score}$\uparrow$ & \textbf{Self-incl.}$\uparrow$ & \textbf{F}$\uparrow$ & \textbf{T}$\uparrow$ & \textbf{H}$\uparrow$ & \textbf{Generator} \\
    \midrule
     1 & Gemini-4-argon        & 0.875 & 0.875 & 0.95 & 0.92 & 0.75 & ---        \\
     2 & Claude-opus-5.5       & 0.853 & 0.853 & 0.97 & 0.92 & 0.67 & ---        \\
     3 & Gemini-3.8-flash      & 0.846 & 0.887 & 0.92 & 0.92 & 0.70 & \checkmark \\
     4 & GPT-6-sol             & 0.837 & 0.837 & 0.94 & 0.96 & 0.61 & ---        \\
     5 & GPT-5.6-sol           & 0.832 & 0.857 & 0.92 & 0.96 & 0.62 & \checkmark \\
     6 & Claude-opus-5         & 0.815 & 0.859 & 0.92 & 0.89 & 0.63 & \checkmark \\
     7 & GPT-6-luna            & 0.783 & 0.783 & 0.91 & 0.95 & 0.49 & ---        \\
     8 & Gemini-3.5-flash-lite & 0.747 & 0.747 & 0.88 & 0.91 & 0.46 & ---        \\
     9 & DeepSeek-V4-Pro       & 0.725 & 0.747 & 0.78 & 0.93 & 0.47 & \checkmark \\
    10 & Gemma-4-31B        & 0.712 & 0.729 & 0.86 & 0.88 & 0.40 & \checkmark \\
    11 & Qwen3.5-397B     & 0.671 & 0.685 & 0.75 & 0.90 & 0.36 & \checkmark \\
    12 & Claude-haiku-5.5      & 0.651 & 0.651 & 0.72 & 0.89 & 0.35 & ---        \\
    13 & GLM-5.3-flash         & 0.646 & 0.672 & 0.74 & 0.80 & 0.40 & \checkmark \\
    14 & Llama-4-Maverick      & 0.609 & 0.631 & 0.69 & 0.82 & 0.32 & \checkmark \\
    15 & Grok-4.3              & 0.547 & 0.576 & 0.55 & 0.81 & 0.28 & \checkmark \\
    16 & Gemma-SEA-LION-9B  & 0.349 & 0.349 & 0.46 & 0.57 & 0.02 & ---        \\
    17 & Sailor2-8B       & 0.348 & 0.348 & 0.43 & 0.50 & 0.12 & ---        \\
    18 & SeaLLMs-v3-7B         & 0.303 & 0.303 & 0.41 & 0.49 & 0.01 & ---        \\
    19 & Llama-SEA-LION-8B  & 0.280 & 0.280 & 0.38 & 0.44 & 0.02 & ---        \\
    \bottomrule
  \end{tabular}
  \caption{Scores averaged over 50 questions and ten trials (Table~\ref{tab:models}). Score, F, T, and H use self-excluded references; the Self-incl. column gives Score with the evaluated model's references retained. Ranks use unrounded scores.}
  \label{tab:leaderboard}
\end{table*}

\paragraph{Allocation.}
\label{app:allocation}
For each $(N, k)$, family combinations are built greedily, adding at each step the family that most increases coverage of the development gold answers; each sampled set is compared with the full pool on the same questions.
The fixed-family analysis uses Claude, DeepSeek, Gemini, and GLM on 39 common questions with 20 random seeds, against a baseline of all references from these four families (188 per question on average).

\paragraph{Family replacement.}
\label{app:replacement}
The baseline draws ten references from each of the target family and three other families.
The replacement condition substitutes an outside family for the target family, and the control substitutes the same outside family for one of the other three.
Score changes are averaged over up to 20 random seeds within each question.

\clearpage
\raggedbottom
\section{Human Evaluation}
\label{app:human}

\paragraph{Response ratings.}
The native Khmer-speaking author rates the 240 test responses with the rubric in Table~\ref{tab:human-items}, after calibrating on 40 development items.
Responses are shown in random order within each question, with model identities, gold answers, references, and benchmark scores hidden; ties are allowed in the ranking, and \emph{unsure} correctness judgments are treated as missing.

\begin{table}[!ht]
  \centering
  \small
  \setlength{\tabcolsep}{4pt}
  \begin{tabular}{@{}p{0.3\columnwidth}lp{0.53\columnwidth}@{}}
    \toprule
    \textbf{Item} & \textbf{Axis} & \textbf{Scale} \\
    \midrule
    Is the Khmer well formed?   & F     & 0: broken \newline 1: readable but unnatural or with errors \newline 2: natural \\
    \addlinespace
    Is the content correct?      & T     & 0: errors or fabrication \newline 1: imprecise details \newline 2: correct (``unsure'' allowed, treated as missing) \\
    \addlinespace
    Does it answer the question? & H     & 0: no \newline 1: partly \newline 2: contains the needed information \\
    \addlinespace
    Overall                      & Score & Rank among the six answers to the question \newline (1 = best; ties allowed) \\
    \bottomrule
  \end{tabular}
  \caption{Human rating criteria and corresponding scorer axes.}
  \label{tab:human-items}
\end{table}

\paragraph{Reference-validity audits.}
The audits use 20 retained references per condition for Japanese and, for Khmer, 50 per Gemma variant, 26 for base Qwen, and 40 for post-trained Qwen.

\end{document}